\documentclass{article} 
\usepackage{iclr2023_conference,times}

\usepackage{amsmath,amsfonts,bm}

\def\eqref#1{equation~\ref{#1}}

\def\1{\bm{1}}

\DeclareMathAlphabet{\mathsfit}{\encodingdefault}{\sfdefault}{m}{sl}
\SetMathAlphabet{\mathsfit}{bold}{\encodingdefault}{\sfdefault}{bx}{n}

\usepackage{multirow}
\usepackage{hyperref}
\usepackage{url}
\usepackage{graphicx}
\usepackage{amsmath}
\usepackage{array}
\usepackage{wrapfig}

\title{ClusTR: Exploring Efficient Self-attention via Clustering for Vision Transformers}
\author{Yutong Xie$^{1}$ \& Jianpeng Zhang$^{2}$ \& Yong Xia$^{3}$ \& Anton van den Hengel$^{1}$ \& Qi Wu$^{1}$\thanks{Corresponding author. }
\\
$^{1}$ The University of Adelaide. \\ $^{2}$ DAMO Academy, Alibaba Group. \\  $^{3}$ Northwestern Polytechnical University. \\
\texttt{\{yutong.xie678\}@gmail.com}; \texttt{\{qi.wu01\}@adelaide.edu.au} \\
}

\begin{document}

\maketitle

\begin{abstract}
Although Transformers have successfully transitioned from their language modelling origins to image-based applications, their quadratic computational complexity remains a challenge, particularly for dense prediction. 
%
In this paper we propose a content-based sparse attention method, as an alternative to dense self-attention, aiming to reduce the computation complexity while retaining the ability to model long-range dependencies. 
Specifically, we cluster and then aggregate key and value tokens, as a content-based method of reducing the total token count. 
The resulting clustered-token sequence retains the semantic diversity of the original signal, but can be processed at a lower computational cost.
Besides, we further extend the clustering-guided attention from single-scale to multi-scale, which is conducive to dense prediction tasks. 
We label the proposed Transformer architecture ClusTR, and demonstrate that it achieves state-of-the-art performance on various vision tasks but at lower computational cost and with 
fewer parameters. 
For instance, our ClusTR small model with 22.7M parameters achieves 83.2\% Top-1 accuracy on ImageNet. 
Source code and ImageNet models will be made publicly available. 
\end{abstract}

\section{Introduction}

Transformers have driven rapid progress in natural language processing, and have become the predominant model in the field as a result~\citep{vaswani2017attention,GPT3}. 
The first Transformer to achieve image recognition performance comparable to the firmly established CNN models (\eg ResNet~\citep{he2016deep} and EfficientNet~\citep{tan2019efficientnet}) was ViT~\citep{ViT}. ViT splits images into $16\times16$ patches, resulting in a sequence of visual tokens. In contrast to the local receptive fields of CNNs, each token in ViT is able to interact with every other token, irrespective of location, thus enabling the modelling of long-range dependencies. 

\begin{wrapfigure}{r}[0cm]{0pt}
\includegraphics[width=0.55\linewidth]{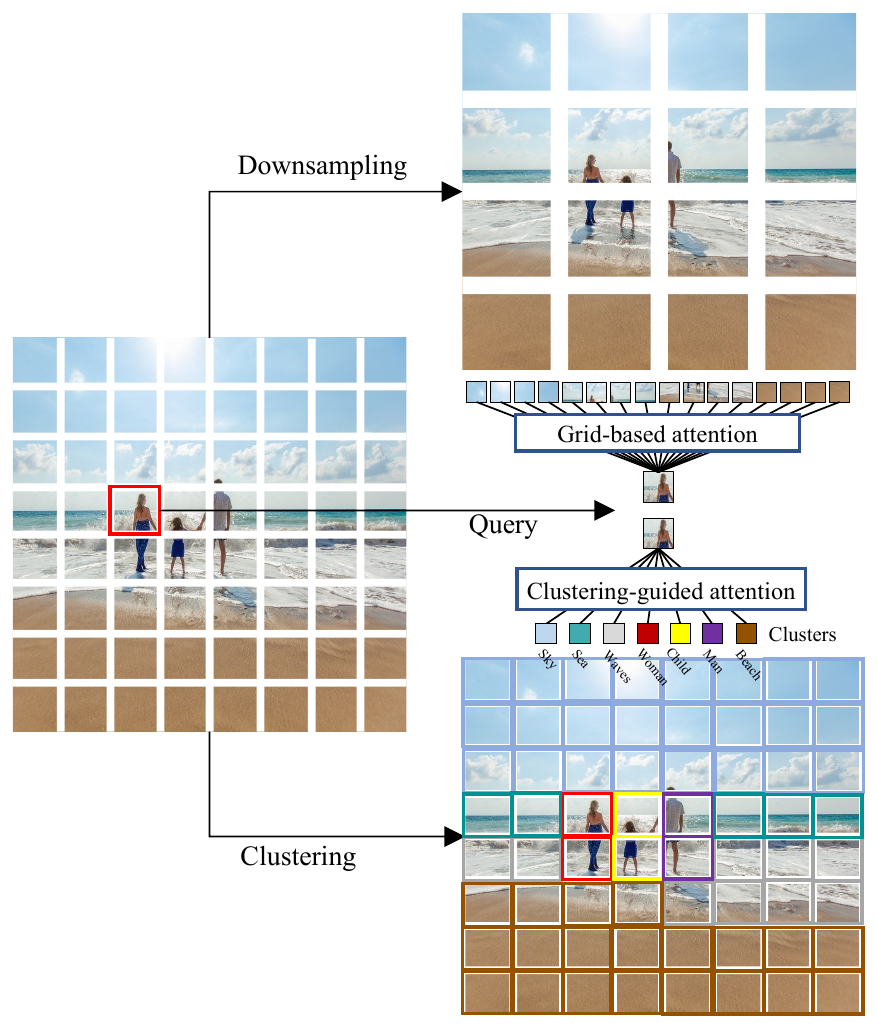}
\caption{Comparison of grid-based self-attention and our clustering-guided self-attention. 
The downsampled grid tokens not only weaken the fine-grained information but also suffer from the redundant information caused by the background or large-size objects. 
With the clustering, the reduced tokens are endowed with explicit and intensive semantics. The large-size object like ``sky'' only occupies one of the entire reduced sequences. Other informative tokens can be assigned with rich and diverse semantic information. 
}
\label{fig1}
\end{wrapfigure}

Although its strength has been demonstrated in various tasks, ViT still suffers from the quadratic complexity in both computation and memory due to the dense token-to-token self-attention. This particularly hinders the applications in dense prediction, such as semantic segmentation. 
Inspired by CNN models~\citep{AlexNet,GoogleNet,he2016deep}, recent research~\citep{Swin,PVT,PiT,chu2021twins} has developed pyramid architectures for Transformers.  
The resultant variation in regulable token length and number of channels at various locations and scales enables greater computational and memory efficiency. 
To further reduce complexity, 
Swin Transformer~\citep{Swin} limited self-attention to a local window, and enabled cross-window connection through the window shifting. 
This means the computational burden scales linearly with the number of tokens, but at the cost of long-range dependencies.
Pyramid Vision Transformer (PVT)~\citep{PVT} reduced the spatial dimension of queries and keys using the large-kernel and large-stride convolution. Such a spatial reduction attention suffers from the following two drawbacks.
First, the reduced tokens are limited by the lack of fine-grained information. As shown in Figure~\ref{fig1}, the downsampled token includes a wide range of content information. Taking the token located in the second row and second column for example, the object of ``woman" only occupies a small part of the whole token, and the token also contains a small part of ``child'' and a large object of ``sky''. This may lead to ambiguous semantics for these tokens.
Second, the background tokens, like the sky and beach, take up quite a large portion of the entire sequence, which are full of redundant information whilst investing most of the computations. 
Hence, the aforementioned deficiencies may have a negative effect on the performance. 

In this paper, we propose a content-based sparse attention to facilitate the efficient and versatile vision Transformer. 
We aim to reduce the self-attention complexity by cutting down the key and value tokens. Different from the grid-based downsampling solution used by~\citep{PVT,wang2021pvtv2}, shown in Figure~\ref{fig1}, we propose to cluster the tokens according to their content similarity and aggregate the tokens in the same cluster as one representative token. This offers the following merits for the vision domain: the clustered tokens contain not only rich but also explicit semantic information, not affected by the background or other large-size objects; and the number of reduced tokens can be set more flexibly than the grid downsampling. 
We then utilize the query tokens, clustered key and value tokens to perform the self-attention, called clustering-guided self-attention. 
Moreover, multi-scale information plays a critical role in vision tasks~\citep{zhang2021multi,chen2021crossvit,ren2022shunted}. We can easily obtain the multi-scale aggregation tokens by adjusting the number of clusters, and further introduce a multi-scale self-attention to strengthen the multi-scale long-range dependencies. 
Since the clustering operation was used, we name the proposed Transformer as ClusTR for clarity. 
We scale up the ClusTR to three variants, ClusTR-T/S/B, corresponding to the tiny, small, and base models. 

We demonstrate the effectiveness of our clustering-based self-attention on several extensive tasks, including classification, segmentation, detection, and pose estimation. The experimental results show that our ClusTR outperforms the competitive CNN-based and Transformer-based counterparts. For instance, ClusTR achieves the 83.2\% and 84.1\% Top-1 accuracy on ImageNet with the 22.7M and 40.3M parameters, respectively. 
Our contributions are summarized as follows:
\begin{itemize}
  \item We propose a content-based self-attention Transformer that clusters and aggregates the visual tokens according to their semantic information. Our clustering-guided self-attention not only maintains the ability for long-range context modelling but also reduces the quadratic computation complexity.
  \item We introduce the multi-scale modelling to our clustering-guided self-attention, leading to the multi-scale self-attention that brings benefits to various dense prediction tasks. 
  \item Our ClusTR, as a versatile Transformer backbone, achieves state-of-the-art performance on four representative vision tasks including classification, segmentation, detection, and pose estimation, setting a new state of the art.
\end{itemize}

\section{Related work}
\label{Related work}

\textbf{Vision Transformer.}
%
%
Transformer, a dominant architecture in language modelling, has recently been extended to the field of computer vision. 
\citep{ViT} designed a vision Transformer (ViT) and successfully achieved comparable or even superior performance on image recognition tasks than the competitive CNN counterparts. Subsequently, many attempts have been made to explore the potential of Transformers in various vision tasks, including segmentation~\citep{SETR}, detection~\citep{DETR}, low-level vision~\citep{chen2021pre}, and image generation~\citep{jiang2021transgan}. 
It is well known that ViT is heavily dependent on a huge amount of data due to its weak inductive bias~\citep{d2021convit}. 
DeiT~\citep{DeiT} utilizes an efficient Transformer optimization strategy that distils another strong classifier to reduce data consumption. 
T2T-ViT\citep{T2T-ViT} models the local image structure via a Tokens-to-Token (T2T) transformation.  
CaiT~\citep{CaiT} uses layer scaling to increase the stability of the optimization when training large-scale Transformers. 
Although achieving record performance on ImageNet~\citep{deng2009imagenet}, these methods suffer from the quadratic complexity of dense self-attention, which is inferior to CNN models when carrying out dense prediction tasks or processing high-resolution images. 
Inspired by CNN models, the community has sprouted the pyramid Transformer structure~\citep{PVT,Swin,PiT,LocalViT,chu2021twins,chen2022regionvit,ren2022shunted}, which breaks with the established patterns in ViT, such as the fixed token length and fixed channels. 
These methods have a pyramid structure that can be used as a versatile backbone for both image classification and dense prediction tasks. 
Among these pyramid Transformer variants, downsampling tokens at each stage is a common but essential operation, which is implemented by convolution with strides~\citep{PVT,wang2021pvtv2}, patch merging with linear projections~\citep{Swin}, or clustering-based patch embedding~\citep{TCFormer}. 

\textbf{Efficient sparse self-attention.}
Different from language models, vision Transformers usually accept a mass of visual tokens, resulting in high computational complexity when computing the dense self-attention, especially on high-resolution images. 
Efficient sparse self-attention, an alternative to vanilla dense self-attention, allows for arbitrary sparsity patterns instead of interacting with all tokens in a dense manner. 
It can be categorized into location-based and content-based methods. 
The location-based sparse attention assumes that not all tokens are valuable and thus only interact with a portion of the tokens. 
Typical examples in this category include local window sliding attention, global attention, and the combination ones~\citep{Beltagy2020Longformer,zaheer2020big}, which have been widely explored in language modelling. 
In the vision field, \citep{Swin} achieved efficient self-attention by limiting self-attention in a local region and interacting with regions through shifting windows. 
\citep{PVT} reduced the key and value tokens by aggregating the local region to a single token through convolution with large kernels and large strides. 
%
However, these artificially-designed sparsity patterns do not necessarily match the characteristics of data, thereby possibly impacting its performance. 
To address this issue, the content-based sparse attention partitions the tokens according to their content correlation. Based on it, \citep{routingTransformer} clustered the tokens using the $k$-means algorithm and performed the self-attention in each cluster. \citep{Reformer} presented an efficient locality sensitivity hashing clustering to divide tokens into chunks. 
\citep{wang2022kvt} proposed the $k$NN attention to select the top-$k$ tokens from keys and ignored the rest for each query when computing the attention matrix, thus filtering out noisy tokens and speeding up training. 
Although spare attention has been studied in these attempts, our ClusTR is different in the following aspects:
1) Compared with \citet{PVT,Swin}, ClusTR breaks the rigid rules of grid-based token aggregation and makes full use of token representation for efficient vision modelling. 
2) \citep{routingTransformer,Reformer,wang2022kvt} limited the range of self-attention to achieve efficiency, in which only similar tokens in the same cluster can communicate with each other. In contrast, our ClusTR breaks the constraints of limited self-attention range, and encourages to explore global attention patterns from the diverse clustered tokens. 
3) Moreover, with the proposed multi-scale attention, ClusTR is superior to these single-scale attention methods when processing dense prediction tasks. 

\section{Method}
\begin{figure}[h]
\centering
\includegraphics[width=1.0\linewidth]{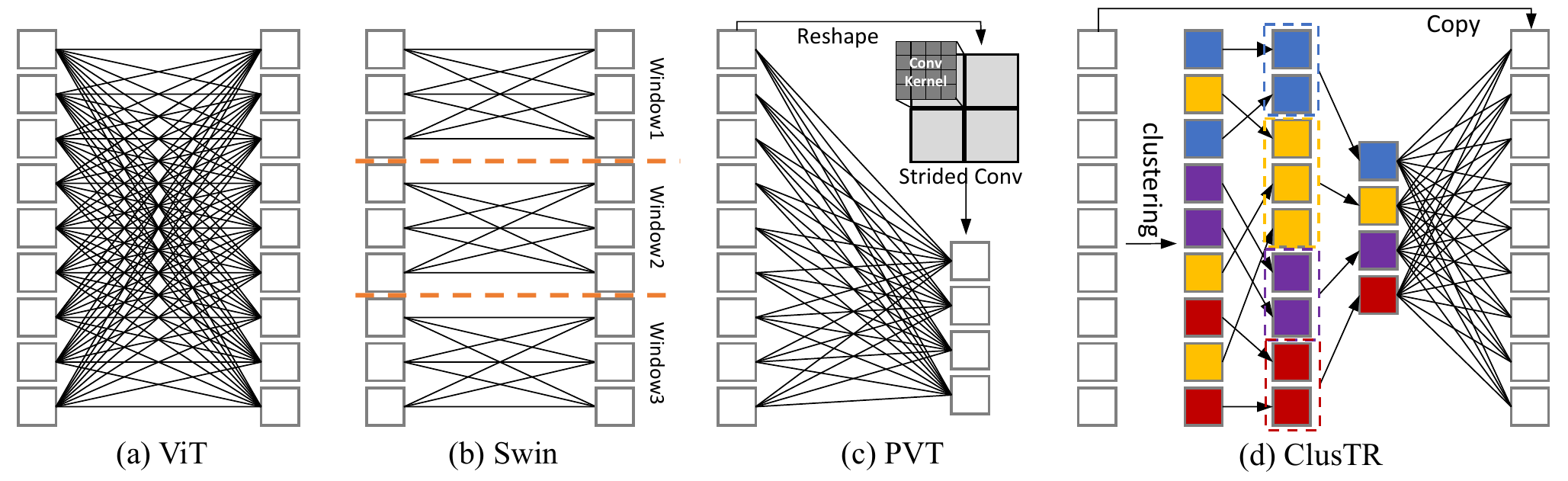}
\caption{Comparison of self-attention in ViT, Swin, PVT and our proposed method. ViT performs the dense token-to-token self-attention; Swin Transformer divides all tokens into several windows and performs the window-based self-attention; PVT aggregates tokens in a grid by using strided convolution. The proposed method groups vision tokens according to the feature similarity, resulting in compact but semantic tokens for efficient self-attention. }
\label{comparison_of_self_attention}
\end{figure}

As an efficient vision Transformer, ClusTR is different from other counterparts in terms of the self-attention mechanism. 
As shown in Figure~\ref{comparison_of_self_attention}, we group vision tokens and aggregate the semantic-similar tokens in the same cluster, aiming to reduce the computational complexity of self-attention. 
Based on the clustering-guided self-attention, we can easily extend it to a multi-scale version which is benefited from the multi-scale aggregation. 
In the following, we delve into the ClusTR self-attention and architecture details.

\subsection{$k$NN-based Density Peaks Clustering}
\label{clustering}
We denote the set of vision tokens as $X=[x_1, x_2, ..., x_N]^\mathsf{T} \in \mathbb{R}^{N\times C}$, where $N$ and $C$ represent the number of tokens and dimension of the token channel, respectively. 
Following~\citep{rodriguez2014clustering}, we characterize token clusters by a higher density than their neighbors and by a relatively large distance from other tokens with higher densities. 
As for a token $x_i\in X$, its local density is defined as 
\begin{equation}
    \rho_i = \exp{(- \frac{1}{k} \sum_{j\in k{\rm NN}(x_i)} d(x_i,x_j)^2)}
\end{equation}
where 
$d(x_i, x_j)$ refers to the Euclidean distance between $x_i$ and $x_j$, 
$k{\rm NN}(x_i) = \{ j\in X | d(x_i,x_j)\leq d(x_i, x_k)\}$, 
$x_k$ is the $k$-th neighbor of $x_i$. 
Here, we also define another variable $\delta_i$ for the token $x_i$, which measures the distance between $x_i$ and other high-density tokens. 
\begin{equation}
    \delta_i = \left\{\begin{matrix}
\underset{j: \rho_j > \rho_i}{\min}(d(x_i, x_j)) & if \; \exists \rho_j>\rho_i\\ 
\underset{j}{\max}(d(x_i, x_j)) & if \; \exists\mkern-10mu/ \rho_j>\rho_i
\end{matrix}\right.
\end{equation}
If $x_i$ is characterized as a cluster, its local density should be higher than that of its neighbors. Besides, it should also have a relatively large distance from other higher-density tokens. To this end, a decision value 
$\gamma_i = \rho_i * \delta_i$ can be computed to locate the density peaks efficiently. The token clusters are specialized with both large density $\rho$ and large distance $\delta$. 
After that, the remaining tokens are assigned to the same cluster as their nearest tokens with higher density. 
Based on the cluster index, we can partition all tokens in $X$ into $M$ clusters, denoted by $G = \{G_1, G_2, ..., G_M\}$. 

The tokens in the same cluster are aggregated to generate a cluster representative token, formulated by
\begin{equation}
\label{eq.clusters}
    \texttt{Cluster}(X;\lambda) = [h_1, h_2, ..., h_{N/\lambda}] \in \mathbb{R}^{M\times C}
\end{equation}
where $\lambda = N/M$ is the token reduction ratio, $h_i = \sum_{x_i \in G_i}^{} w_i \cdot x_i$, and $w_i$ is the learnable parameter for each token $x_i$. 
Note that the number of aggregated cluster representative tokens is far smaller than that of the original visual tokens $X$, \textit{i.e.}, $N>>N/\lambda$. 
Such a clustering-guided token aggregation condenses a lot of visual tokens, benefiting the efficient self-attention process.

\subsection{Clustering-Guided Self-Attention}
The attention module is one of the core components of the Transformer.  Following~\citep{vaswani2017attention}, most of Transformers and their variants apply the multi-head self-attention mechanism to model the long-range dependencies. 
For each head, the query $Q$, key $K$, and value $V$ have the size of $N\times C$. 
The scaled dot-product attention can be formulated as
\begin{equation}
    \texttt{Attention}(Q, K, V) = {\rm Softmax}(\frac{Q K^\mathsf{T}}{\sqrt{s}}) V
\end{equation}
where $s$ is the scaling factor. 
Although the above self-attention can be implemented in a fast manner by using highly optimized matrix multiplication, it still suffers from the high computation complexity, \textit{i.e.}, $O(N^2)$, especially for the abundant vision tokens. 
To address this issue, we propose a clustering-guided efficient self-attention that clusters and aggregates the semantic-similar tokens in the same cluster to reduce the computation complexity. Based on the clustering algorithm in Sec.~\ref{clustering}, the proposed efficient self-attention is reformulated as 
\begin{equation}
\label{eq.self_attention}
    \texttt{ClusAtt}(Q, K, V; \lambda) = {\rm Softmax}(\frac{Q \cdot \texttt{Cluster}(K;\lambda)^\mathsf{T}}{\sqrt{s}}) \cdot \texttt{Cluster}(V;\lambda)
\end{equation}
After clustering, the tokens of key and value are decreased by $\lambda$ times, reducing the computation complexity from $O(N^2)$ to $O(\frac{N^2}{\lambda})$. Based on the single-head attention, the multi-head attention can be implemented in parallel as
\begin{equation}
    \texttt{MH-ClusAtt}(X; \lambda) = \Phi( \bigcup_{i=1}^{H} \texttt{ClusAtt}(XW_i^Q, XW_i^K, XW_i^V; \lambda))
\end{equation}
where $\cup$ refers to the concatenation operation,
$\Phi$ aggregates the feature representation of $H$ attention heads through a linear projection function. $W_i^Q$, $W_i^K$, and $W_i^V$ are linear projections to generate query, key, and value tokens.

\subsection{Multi-Scale Self-Attention}
Here we extend the proposed clustering-guided self-attention from single-scale to multi-scale. 
For the multi-scale aggregation, we replace the single $\lambda$ in Eq.~\ref{eq.clusters} with a set of factors ${\lambda_1, ..., \lambda_{L}}$, where $L$ refers to the number of scales. Then, the multi-scale clustering can be described as 
\begin{equation}
    \texttt{Cluster}(X;{\lambda_1, ..., \lambda_{L}}) = [h_1^{\lambda_1}, ..., h_{N/\lambda_1}^{\lambda_1}; ...; h_1^{\lambda_{L}}, ..., h_{N/\lambda_{L}}^{\lambda_{L}}]  \in \mathbb{R}^{(\frac{N}{\lambda_1}+...+\frac{N}{\lambda_{L}})\times C}
\end{equation}
The computational complexity of multi-scale attention is $O(N^2(\frac{1}{\lambda_1}+...+\frac{1}{\lambda_{L}}))$. And the multi-head multi-scale clustering-guided self-attention can be described as
\begin{equation}
    \texttt{MHMS-ClusAtt}(X;{\lambda_1, ..., \lambda_{L}}) = \Phi( \sum_{j=1}^{L} \bigcup_{i=1}^{H} \texttt{ClusAtt}(XW_i^Q, XW_i^K, XW_i^V; \lambda_j))
\end{equation}
where the linear projection $\Phi$ is used to aggregate the feature representation of $H$ attention heads and $L$ scales.

\subsection{ClusTR Transformer Architecture}
\begin{figure}[!t]
\centering
\includegraphics[width=1.0\linewidth]{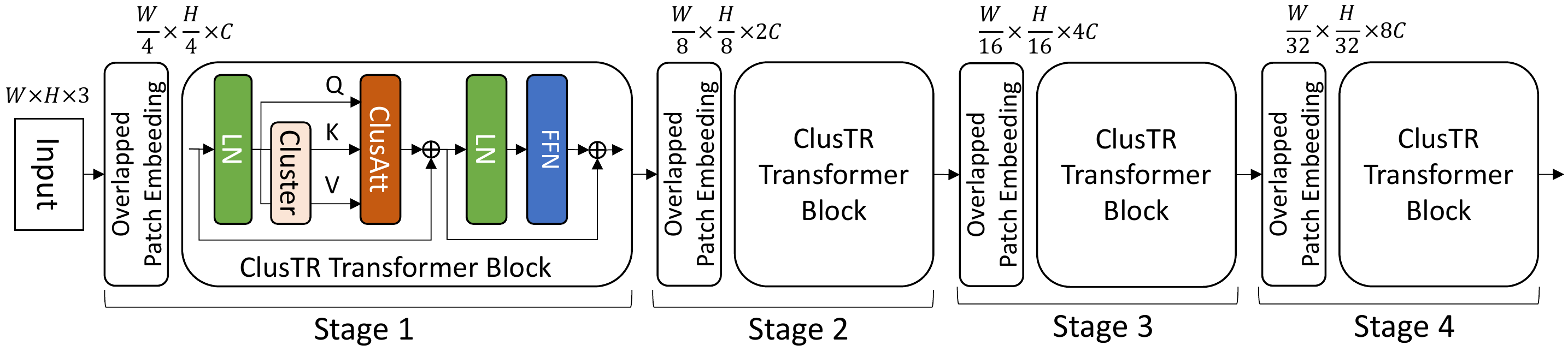}
\caption{The architecture of our ClusTR. }
\label{ClusTR_architecture}
\end{figure}

The basic ClusTR model is composed of four stages, as shown in Figure~\ref{ClusTR_architecture}. 
We follow ~\citep{ren2022shunted} and employ the overlapped patch embedding at the beginning of each stage to model local continuity. 
%
Based on the clustering-guided self-attention, the Transformer block of ClusTR can be computed as
\begin{equation}
\begin{split}
  &z_l' = \texttt{MHMS-ClusAtt}(\texttt{LN}(z_{l-1})) + z_{l-1} \\
  &z_l = \texttt{FFN}(\texttt{LN}(z_l')) + z_l' \\
\end{split}
\end{equation}
where $\texttt{LN}$ is the layer normalization, and $\texttt{FFN}$ is the fully connected feedforward network.
%
Note that the token reduction ratio $\lambda$ can be defined as any value during the clustering process. 
To balance the efficiency and accuracy, we set $\lambda$ to $\{64,16\}$, $\{16,4\}$, $\{4,1\}$, $\{1\}$ from the first to the last stage, respectively. 
We build the tiny model, called ClusTR-T, that has a similar model size and computation complexity to PVT-Tiny/PVTv2-B1. Based on this, we scale up ClusTR-T to the small and base variants, called ClusTR-S, and ClusTR-B, which have the model size and computation complexity of about $2\times$, and $4\times$ compared to the tiny version. 
The specific architecture details and hyper-parameters can be found in Table~\ref{tab:tab0}. 

\begin{table}[t]
\small
\caption{Architecture details of ClusTR variants. Here `L, C, H' represents the number of Transformer layers, channels, and heads, respectively.}
\label{tab:tab0}
\centering
\begin{tabular}{c|c|m{0.2cm}<{\centering}m{0.2cm}<{\centering}m{0.2cm}<{\centering}m{0.8cm}<{\centering}|m{0.2cm}<{\centering}m{0.2cm}<{\centering}m{0.2cm}<{\centering}m{0.8cm}<{\centering}|m{0.2cm}<{\centering}m{0.2cm}<{\centering}m{0.2cm}<{\centering}m{0.8cm}<{\centering}}
\hline
\multirow{2}{*}{} & \multirow{2}{*}{Output\_size} & \multicolumn{4}{c|}{ClusTR-T}                                                            & \multicolumn{4}{c|}{ClusTR-S}                                                            & \multicolumn{4}{c}{ClusTR-B}                                                             \\ \cline{3-14} 
                  &                               & \multicolumn{1}{c|}{L} & \multicolumn{1}{c|}{C}   & \multicolumn{1}{c|}{H} & $\lambda$       & \multicolumn{1}{c|}{L}  & \multicolumn{1}{c|}{C}   & \multicolumn{1}{c|}{H} & $\lambda$       & \multicolumn{1}{c|}{L}  & \multicolumn{1}{c|}{C}   & \multicolumn{1}{c|}{H} & $\lambda$       \\ \hline
Stage1            & W/4 * H/4                     & \multicolumn{1}{c|}{1} & \multicolumn{1}{c|}{64}  & \multicolumn{1}{c|}{1} & \{64,16\} & \multicolumn{1}{c|}{3}  & \multicolumn{1}{c|}{64}  & \multicolumn{1}{c|}{1} & \{64,16\} & \multicolumn{1}{c|}{3}  & \multicolumn{1}{c|}{64}  & \multicolumn{1}{c|}{1} & \{64,16\} \\ \hline
Stage2            & W/8 * H/8                     & \multicolumn{1}{c|}{2} & \multicolumn{1}{c|}{128} & \multicolumn{1}{c|}{2} & \{16,4\} & \multicolumn{1}{c|}{5}  & \multicolumn{1}{c|}{128} & \multicolumn{1}{c|}{2} & \{16,4\} & \multicolumn{1}{c|}{5}  & \multicolumn{1}{c|}{128} & \multicolumn{1}{c|}{2} & \{16,4\} \\ \hline
Stage3            & W/16 * H/16                   & \multicolumn{1}{c|}{6} & \multicolumn{1}{c|}{256} & \multicolumn{1}{c|}{4} & \{4,1\} & \multicolumn{1}{c|}{13} & \multicolumn{1}{c|}{256} & \multicolumn{1}{c|}{4} & \{4,1\} & \multicolumn{1}{c|}{18} & \multicolumn{1}{c|}{320} & \multicolumn{1}{c|}{5} & \{4,1\} \\ \hline
Stage4            & W/32 * H/32                   & \multicolumn{1}{c|}{1} & \multicolumn{1}{c|}{512} & \multicolumn{1}{c|}{8} & 1       & \multicolumn{1}{c|}{2}  & \multicolumn{1}{c|}{512} & \multicolumn{1}{c|}{8} & 1       & \multicolumn{1}{c|}{3}  & \multicolumn{1}{c|}{512} & \multicolumn{1}{c|}{8} & 1       \\ \hline
\end{tabular}
\end{table}

\section{Experiment}
We evaluate ClusTR on four representative computer vision tasks, including image classification, semantic segmentation, object detection, and pose estimation.
We also investigate the effectiveness of each part of ClusTR in the ablation section. 

\subsection{Classification on ImageNet-1K}
\textbf{Dataset: }
We conduct image classification experiments on the ImageNet-1K dataset~\citep{deng2009imagenet}, which includes 1.28 million training images and 50K validation images from 1,000 categories.

\textbf{Setting: }
We randomly crop $224\times224$ regions as the input. Following~\citep{wang2021pvtv2}, we apply a rich set of data augmentations to diversify the training set, including random cropping, random flipping, random erasing, label-smoothing regularization, CutMix, and Mixup. We adopt the AdamW optimizer~\citep{adamw} with a cosine decaying learning rate~\citep{cosine_LR}, a momentum of 0.9, and a weight decay of 0.05, to train our ClusTR model. We set the initial learning rate to 0.001, batch size to 1024, and epochs to 300, which are popular for ImageNet training. 
During the inference time, we take a $224\times224$ center crop as the input and adapt the Top-1 accuracy as the evaluation metric.

\textbf{Results: }
In Table~\ref{tab:tab1}, we compare ClusTR to other advanced backbones based on ConvNets, MLPs, and Transformers. 
Compared with the state-of-the-art Transformer-based methods, 
ClusTR outperforms the Transformer-based architectures with comparable or fewer parameters and computation budgets, surpassing 1.9\% than Swin Transformer (ClusTR-S 83.2 vs. Swin-T 81.3), and 1.5\% than PVTv2 (ClusTR-T 80.2 vs. PVTv2-b1 78.7). 
Compared to the state-of-the-art ConvNet-based methods, 
ClusTR is superior to keep a balance between accuracy and complexity. 
With a similar complexity budget, ClusTR achieves 1.1\% performance gain over ConvNets (ClusTR-S 83.2 vs. ConvNeXt-T 82.1). 
With a comparable accuracy (ClusTR 83.2 vs. ConvNeXt-S 83.1), ClusTR reduces the model complexity of ConvNexts by half (ClusTR-S 22.7M/4.8G vs. ConvNeXt-S 50M/8.7G). 
Such an advantageous accuracy-complexity trade-off still remains when compared to MLP-based methods.

%
%

\begin{table}[t]
\small
\caption{Image classification performance of different backbones on the ImageNet-1K validation set. Here `Params.' refers to the number of the model parameters, and FLOPs is calculated based on the input size of $224\times 224$. }
\label{tab:tab1}
\centering
\begin{tabular}{cccccc}
\hline
Methods       & Resolution & \begin{tabular}[c]{@{}c@{}}Params. \\ (M)\end{tabular} & FLOPs (G) & Top-1 (\%) & Reference \\ \hline
\multicolumn{6}{c}{ConvNets}                                                  \\ \hline
RegNetY-4G~\citep{radosavovic2020designing}    & 224        & 21.0        & 4.0       & 80.0      & CVPR20    \\
RegNetY-8G~\citep{radosavovic2020designing}    & 224        & 39.0        & 8.0       & 81.7      & CVPR20    \\
ConvNeXt-T~\citep{liu2022convnet}    & 224        & 29.0        & 4.5       & 82.1      & CVPR22    \\
ConvNeXt-S~\citep{liu2022convnet}    & 224        & 50.0        & 8.7       & 83.1      & CVPR22    \\ \hline
\multicolumn{6}{c}{MLPs}                                                \\ \hline
CycleMLP-T~\citep{chen2022cyclemlp}    & 224        & 28.0        & 4.4       & 81.3      & ICLR22    \\
CycleMLP-S~\citep{chen2022cyclemlp}    & 224        & 50.0        & 8.5       & 82.9      & ICLR22    \\
AS-MLP-T~\citep{lian2022mlp}      & 224        & 28.0        & 4.4       & 81.3      & ICLR22    \\
AS-MLP-S~\citep{lian2022mlp}      & 224        & 50.0        & 8.5       & 83.1      & ICLR22    \\ \hline
\multicolumn{6}{c}{Transformers}                                        \\ \hline
PVT-T~\citep{PVT}   & 224        & 13.0        & 1.9       & 75.1      & ICCV21    \\
PVT-ACmix-T~\citep{pan2022integration}   & 224        & 13.0        & 2.0       & 78.0      & CVPR22    \\
PVTv2-b1~\citep{wang2021pvtv2}      & 224        & 13.1        & 2.1       & 78.7      & CVM22     \\
QuadTree-B-b1~\citep{tang2022quadtree} & 224        & 13.6        & 2.3       & 80.0      & ICLR22    \\
ClusTR-T          & 224        & 11.7        & 2.2       &  \textbf{80.2}         & Ours \\ \hline
PVT-S~\citep{PVT}   & 224        & 24.5        & 3.8       & 79.8      & ICCV21    \\		
Swin-T~\citep{Swin}        & 224        & 29.0        & 4.5       & 81.3      & ICCV21    \\
Twins-SVT-S~\citep{chu2021twins}   & 224        & 24.0        & 2.9       & 81.7      & NeurIPS21 \\
PVTv2-b2~\citep{wang2021pvtv2}      & 224        & 25.4        & 4.0       & 82.0      & CVM22     \\
HRViT-b2~\citep{gu2022multi}      & 224        & 32.5        & 5.1       & 82.3      & CVPR22    \\
TCFormer~\citep{zeng2022not}      & 224        & 25.6        & 5.9       & 82.4      & CVPR22    \\
CrossFormer-S~\citep{wang2022crossformer} & 224        & 30.7        & 4.9       & 82.5      & ICLR22    \\
RegionViT-S~\citep{chen2022regionvit}   & 224        & 30.6        & 5.3       & 82.6      & ICLR22    \\
CSWin-T~\citep{dong2022cswin}      & 224        & 23.0        & 4.3       & 82.7      & CVPR22    \\ 
QuadTree-B-b2~\citep{tang2022quadtree} & 224        & 24.2        & 4.5       & 82.7      & ICLR22    \\
ClusTR-S          & 224        & 22.7        & 4.8       & \textbf{83.2}          & Ours\\ \hline
PVT-L~\citep{PVT}   & 224        & 61.4	      & 9.8	       & 81.7      & ICCV21    \\
HRViT-b3~\citep{gu2022multi}      & 224        & 37.9        & 5.7       & 82.8      & CVPR22    \\
Swin-S~\citep{Swin}        & 224        & 50.0        & 8.7       & 83.0      & ICCV21    \\
RegionViT-M~\citep{chen2022regionvit}   & 224        & 41.2        & 7.4       & 83.1      & ICLR22    \\
Twins-SVT-B~\citep{chu2021twins}   & 224        & 56.0        & 8.6       & 83.2      & NeurIPS21 \\
CrossFormer-B~\citep{wang2022crossformer}	& 224	   & 52.0 	     & 9.2 	     & 83.4 	 & ICLR22 \\
PVTv2-b4~\citep{wang2021pvtv2}      & 224        & 62.6        & 10.1      & 83.6      & CVM22     \\
Quadtree-B-b3~\citep{tang2022quadtree} & 224        & 46.3        & 7.8       & 83.7      & ICLR22    \\
ClusTR-B          & 224        & 40.2        & 7.5       & \textbf{84.1}          & Ours\\ \hline
\end{tabular}
\end{table}

\subsection{Semantic Segmentation on ADE20K}

\textbf{Dataset: }
We conduct semantic segmentation experiments on the ADE20K dataset~\citep{zhou2017scene}, which includes 20,210 training images and 2,000 validation images from 150 fine-grained semantic categories.

\textbf{Settings: }
We randomly resize and crop $512\times512$ image patches as the input and set the batch size to 16. 
We empoy the ClusTR-S, pre-trained on ImageNet, as the backbone, and evaluate it with two segmentation architectures, \textit{i.e.}, Semantic FPN~\citep{2019Panoptic} and UperNet~\citep{xiao2018unified}. The segmentation training process follows the default settings in ~\citep{wang2021pvtv2} and ~\citep{Swin}. 
When training the Semantic FPN, we adopt the AdamW optimizer~\citep{adamw} with an initial learning rate of 0.0001 and a weight decay of 0.0001, and set the number of iterations to 80K. 
As for UperNet, we adopt the AdamW optimizer with an initial learning rate of 0.00006 and a weight decay of 0.01, and set the number of iterations to 160K. We also warm up the model linearly for the first 1500 iterations. 
During the test, we re-scale the shorter side of the input image to $512$ pixels and adapt the mIOU metric for evaluation.

\textbf{Results: }
As shown in Table~\ref{tab:tab2}, we can see that ClusTR outperforms the state-of-the-art backbones, including ConvNets-based and Transformer-based, in both semantic FPN and UpperNet mode. 
Compared to the ConvNet-based backbones, the proposed ClusTR achieves better segmentation performance (ClusTR 48.0 vs. ResNet 36.7 with Semantic FPN; ClusTR 49.6 vs. ConvNeXt 46.0 with UpperNet) while using fewer parameters. 
Compared with the Transformer-based methods, ClusTR outperforms the state-of-the-art backbones in both semantic FPN and UpperNet mode with comparable or even fewer parameters, surpassing CrossFormer~\citep{wang2022crossformer} by 2.0\%, and MPViT~\citep{Lee_2022_CVPR} by 1.3\%. 


\begin{table}[]
\small
\caption{Semantic segmentation performance of different backbones on the ADE-20K validation set. Here `*' indicates that the numbers are cited from the reproduced results of Twins.}
\label{tab:tab2}
\centering
\begin{tabular}{c|cc|cc}
\hline
                & \multicolumn{2}{c|}{Semantic FPN 80k}       & \multicolumn{2}{c}{UperNet 160K}           \\ \hline
Methods         & \multicolumn{1}{c|}{Params. (M)} & mIOU (\%) & \multicolumn{1}{c|}{Params. (M)} & mIOU (\%) \\ \hline
ResNet-50~\citep{he2016deep}       & \multicolumn{1}{c|}{28.5}       & 36.7      & \multicolumn{1}{c|}{-}          & -         \\ 
PVT-S~\citep{PVT}           & \multicolumn{1}{c|}{28.2}       & 39.8      & \multicolumn{1}{c|}{-}          & -         \\ 
Swin-T*~\citep{Swin}          & \multicolumn{1}{c|}{31.9}       & 41.5      & \multicolumn{1}{c|}{59.9}       & 44.5      \\ 
CycleMLP-b2~\citep{chen2022cyclemlp}     & \multicolumn{1}{c|}{30.6}       & 43.4      & \multicolumn{1}{c|}{-}          & -         \\ 
ConvNeXt-T~\citep{liu2022convnet}      & \multicolumn{1}{c|}{-}          & -         & \multicolumn{1}{c|}{60.0}       & 46.0      \\ 
Twins-SVT-S~\citep{chu2021twins}     & \multicolumn{1}{c|}{28.3}       & 43.2      & \multicolumn{1}{c|}{54.4}       & 46.2      \\ 
RegionViT-S+~\citep{chen2022regionvit}    & \multicolumn{1}{c|}{35.7}       & 45.3      & \multicolumn{1}{c|}{-}          & -         \\ 
CrossFormer-S~\citep{wang2022crossformer}   & \multicolumn{1}{c|}{34.3}       & 46.0      & \multicolumn{1}{c|}{62.3}       & 47.6      \\ 
MPViT-S~\citep{Lee_2022_CVPR}         & \multicolumn{1}{c|}{-}          & -         & \multicolumn{1}{c|}{52.0}       & 48.3      \\ 
ClusTR-S (Ours) & \multicolumn{1}{c|}{26.4}       & \textbf{48.0}       & \multicolumn{1}{c|}{52.5}       & \textbf{49.6}      \\ \hline
\end{tabular}
\end{table}

\subsection{Object detection on COCO}
\textbf{Dataset: }
We perform object detection and instance segmentation experiments on the COCO2017 dataset~\citep{lin2014microsoft}, which includes 118,287 training images and 5,000 validation images from 80 categories.

\textbf{Settings: }
We use the ClusTR-S pre-trained on ImageNet as the backbone of two mainstream detectors, \textit{i.e.}, RetinaNet~\citep{lin2017focal} and Mask R-CNN~\citep{he2017mask}. 
We follow the default settings of PVTv2~\citep{wang2021pvtv2} and mmdetection~\citep{chen2019mmdetection}. We adopt the AdamW optimizer with a batch size of 16, and perform the $1\times$ training schedule with 12 epochs. During training, we re-scale the shorter side of the input image to $800$ pixels while keeping the longer side no more than $1,333$ pixels.
During test, the shorter side of input images is resized to $800$ pixels, and the bbox mAP (AP$^b$) and mask mAP (AP$^m$) are used as evaluation metrics.

\textbf{Results: } 
As shown in Table~\ref{tab:tab3}, with comparable/fewer parameters, our ClusTR model surpasses both ConvNet- and Transformer-based state-of-the-art backbones when using Mask-RCNN for object detection and instance segmentation. 
Compared to ConvNet backbones, our model outperforms ResNet~\citep{he2016deep} by 9.0 points for box AP, and 8.1 points for mask AP. 
Compared to Transformer backbones, our model achieves 6.6 box AP/4.7 mask AP over PVT, and 4.8 box AP/3.4 mask AP over Swin. 
%
Besides, Table~\ref{tab:tab4} reports the detection performance of different backbones when using RetinaNet as a detector. 
Our model achieves the 45.8 box AP with only 32.4M parameters, outperforming other competitors especially in detecting small objects. 
We clarify that these results are expected, since the proposed clustering-guided self-attention is able to pay equal attention to diverse objects, insensitive to the object size, which is particularly beneficial for small objects. 

%

\begin{table}[]
\small
\caption{Detection and instance segmentation performance of Mask-RCNN with different backbones on the COCO validation set.}
\label{tab:tab3}
\centering
\begin{tabular}{c|ccccccc}
\hline
Methods       & Params. (M) & AP$^b$  & AP$^b_{50}$ & AP$^b_{75}$ & AP$^m$  & AP$^m_{50}$ & AP$^m_{75}$ \\ \hline
ResNet-50~\citep{he2016deep}     & 44.2       & 38.0 & 58.6  & 41.4  & 34.4 & 55.1  & 36.7  \\
PVT-S~\citep{PVT}         & 44.1       & 40.4 & 62.9  & 43.8  & 37.8 & 60.1  & 40.3  \\
Swin-T~\citep{Swin}        & 47.8       & 42.2 & 64.6  & 46.2  & 39.1 & 61.6  & 42.0  \\
Twins-SVT-S~\citep{chu2021twins}   & 44.0       & 43.4 & 66.0  & 47.3  & 40.3 & 63.2  & 43.4  \\
CrossFormer-S~\citep{wang2022crossformer} & 50.2       & 45.4 & 68.0  & 49.7  & 41.4 & 64.8  & 44.6  \\
ClusTR-S (Ours)          & 42.3     &       \textbf{47.0}     & \textbf{68.7}     & \textbf{51.6}      & \textbf{42.5}      & \textbf{65.9}     & \textbf{45.9}        \\ \hline
\end{tabular}
\end{table}

\begin{table}[]
\small
\caption{Detection performance of RetinaNet with different backbones on the COCO validation set.}
\label{tab:tab4}
\centering
\begin{tabular}{c|ccccccc}
\hline
Methods       & Params. (M) & AP$^b$  & AP$^b_{50}$ & AP$^b_{75}$ & AP$_S$  & AP$_M$ & AP$_L$ \\ \hline
ResNet-50~\citep{he2016deep}     & 37.7       & 36.3 & 55.3 & 38.6 & 19.3 & 40.0 & 48.8 \\
PVT-S~\citep{PVT}         & 34.2       & 40.4 & 61.3 & 43.0 & 25.0 & 42.9 & 55.7 \\
CycleMLP-b2~\citep{chen2022cyclemlp}   & 36.5       & 40.6 & 61.4 & 43.2 & 22.9 & 44.4 & 54.5 \\
Swin-T~\citep{Swin}        & 38.5       & 41.5 & 62.1 & 44.2 & 25.1 & 44.9 & 55.5 \\
Twins-SVT-S~\citep{chu2021twins}   & 34.3       & 43.0 & 64.2 & 46.3 & 28.0 & 46.4 & 57.5 \\
RegionViT-B~\citep{chen2022regionvit}   & 83.4       & 43.3 & 65.2 & 46.4 & 29.2 & 46.4 & 57.0 \\
CrossFormer-S~\citep{wang2022crossformer} & 40.8       & 44.4 & 65.8 & 47.4 & 28.2 & 48.4 & 59.4 \\
Shunted-S~\citep{ren2022shunted}     & 32.1       & 45.4 & 65.9 & 49.2 & 28.7 & 49.3 & 60.0 \\
ClusTR-S (Ours)          & 32.4          & \textbf{45.8}           & \textbf{66.4}      & \textbf{49.5}           & \textbf{30.4}     & \textbf{49.5}      & \textbf{61.2}      \\ \hline
\end{tabular}
\end{table}

\subsection{2D Whole-body Pose Estimation on COCO.}

\textbf{Dataset: }
We perform pose estimation experiments on the COCOWholeBody V1.0 dataset~\citep{jin2020whole}, which contains 133 keypoints, including 17 for the body, 6 for the feet, 68 for the face, and 42 for the hands. 

\textbf{Implementation details: }
We follow the same settings in~\citep{TCFormer}, and adopt the AdamW optimizer with an initial learning rate of 0.0005~\citep{cosine_LR}, a momentum of 0.9, and a weight decay of 0.01. We set the batch size to 512, and the number of epochs to 210. 
The OKS-based Average Precision (AP) and Average Recall (AR) are used as evaluation metrics. 

\textbf{Results: }
In Table~\ref{tab:tab5}, we compare ClusTR with other advanced models on COCOWholeBody V1.0 dataset. Our model achieves the new state-of-the-art performance on the pose estimation (59.4\% AP and 69.7\% AR), outperforming the best ConvNet-based HRNet by 4.1 AP and 7.1 AR, and surpassing the best Transformer-based TCFormer by 2.2 AP and 1.9 AR.

\begin{table}[]
\scriptsize
\caption{Pose estimation performance of different backbones on the COCOWholeBody V1.0 dataset. Here `*' indicates that the numbers are cited from the reproduced results of TCFormer.}
\label{tab:tab5}
\centering
\begin{tabular}{c|c|cc|cc|cc|cc|cc}
\hline
\multirow{2}{*}{Methods} & \multirow{2}{*}{Resolution}  & \multicolumn{2}{c|}{body}              & \multicolumn{2}{c|}{foot}              & \multicolumn{2}{c|}{face}              & \multicolumn{2}{c|}{hand}              & \multicolumn{2}{c}{whole}             \\ \cline{3-12} 
                       &  & \multicolumn{1}{c|}{AP} & AR & \multicolumn{1}{c|}{AP} & AR & \multicolumn{1}{c|}{AP} & AR & \multicolumn{1}{c|}{AP} & AR & \multicolumn{1}{c|}{AP} & AR \\ \hline
ZoomNet*~\citep{jin2020whole}    &   384$\times$288            & \multicolumn{1}{c|}{74.3}  & 80.2  & \multicolumn{1}{c|}{79.8}  & 86.9  & \multicolumn{1}{c|}{62.3}  & 70.1  & \multicolumn{1}{c|}{40.1}  & 49.8  & \multicolumn{1}{c|}{54.1}  & 65.8  \\ \hline
SBL-Res152*~\citep{xiao2018simple}   &  256$\times$192           & \multicolumn{1}{c|}{68.2}  & 76.4  & \multicolumn{1}{c|}{66.2}  & 78.8  & \multicolumn{1}{c|}{62.4}  & 72.8  & \multicolumn{1}{c|}{48.2}  & 60.6  & \multicolumn{1}{c|}{54.8}  & 66.1  \\ \hline
HRNet-w32*~\citep{sun2019deep}   &   256$\times$192          & \multicolumn{1}{c|}{70.0}  & 74.6  & \multicolumn{1}{c|}{56.7}  & 64.5  & \multicolumn{1}{c|}{63.7}  & 68.8  & \multicolumn{1}{c|}{47.3}  & 54.6  & \multicolumn{1}{c|}{55.3}  & 62.6  \\ \hline
PVTv2-b2~\citep{wang2021pvtv2}   &  256$\times$192            & \multicolumn{1}{c|}{69.6}        & 77.3        & \multicolumn{1}{c|}{69.0}        & 80.3        & \multicolumn{1}{c|}{64.9}        & 74.8        & \multicolumn{1}{c|}{54.5}        & 65.9        & \multicolumn{1}{c|}{57.5}        & 68.0        \\ \hline
TCFormer~\citep{TCFormer}   &  256$\times$192            & \multicolumn{1}{c|}{69.1}  & 77.0  & \multicolumn{1}{c|}{69.8}  & 81.3  & \multicolumn{1}{c|}{64.9}  & 74.6  & \multicolumn{1}{c|}{53.5}  & 65.0  & \multicolumn{1}{c|}{57.2}  & 67.8  \\ \hline
ClusTR-S (Ours)   &  256$\times$192      & \multicolumn{1}{c|}{\textbf{71.4}}        & \textbf{78.8}        & \multicolumn{1}{c|}{\textbf{73.3}}        & \textbf{83.8}        &  \multicolumn{1}{c|}{\textbf{66.5}}        & \textbf{75.7}        &  \multicolumn{1}{c|}{\textbf{55.9}}        & \textbf{67.1}        & \multicolumn{1}{c|}{\textbf{59.4}}        & \textbf{69.7}        \\ \hline
\end{tabular}
\end{table}

\subsection{Ablations}
We perform the following ablation experiments to further verify the effectiveness of ClusTR. All classification experiments are conducted based on ClusTR-T and the number of training epochs is set to 100. 
The segmentation experiments are conducted based on the pre-trained ClusTR-T and the Semantic FPN segmentation architecture.

\textbf{Grid-based vs. clustering-guided token aggregation:}
Token aggregation is an important operation in the self-attention process that dramatically reduces the computation complexity. 
We compare the clustering-guided token aggregation to the convolution-based grid token aggregation. 
Following \citep{PVT}, we utilize the convolution with large strides to achieve the grid token aggregation. Note that the other settings are the same for a fair comparison. 
Table~\ref{tab:tab6} reveals that our clustering-guided method not only reduce the parameters and FLOPs, but also improve 0.5 points of Top1 accuracy (grid-based 76.7 \textit{vs.} clustering-guided 77.2). 


\begin{table}[t!]  
\small
\caption{Comparison of different token aggregation methods on the ImageNet-1K dataset.}
\label{tab:tab6}
\centering
\begin{tabular}{c|c|c|c}
\hline
Token Aggregation & Params. (M) & FLOPs (G) & Top1 (\%) \\ \hline
Grid-based       & 13.2       & 2.1       & 76.7          \\ \hline
Clustering        & 10.8       & 2.0       & 77.2         \\ \hline
\end{tabular}
\end{table}

\textbf{Compared to different sparse attentions:} 
We also compare the clustering-guided attention to the spatial-reduction attention (SRA) as done in PVT~\citep{PVT} and $k$NN based sparse attention~\citep{wang2022kvt} in Table~\ref{tab:tab8}. 
It reveals that the $k$NN attention achieves a slight performance gain (0.1 points) over SRA without increasing parameters and FLOPs. 
It is noteworthy that our ClusTR not only outperforms $k$NN attention by 0.4\% but also reduces about 18\% parameters. It demonstrates that the proposed ClusTR is superior to modelling the token-wise dependencies, thus leading to better performance.

\begin{table}[t!]  
\small
\caption{Comparison of different sparse attentions on the ImageNet-1K dataset.}
\label{tab:tab8}
\centering
\begin{tabular}{c|c|c|c}
\hline
Methods & Params. (M) & FLOPs (G) & Top1 (\%) \\ \hline
Spatial-reduction attention~\citep{PVT}       & 13.2       & 2.1       &  76.7          \\ \hline
$k$NN attention~\citep{wang2022kvt}       & 13.2       & 2.1       &  76.8          \\ \hline
ClusTR (Ours)        & 10.8       & 2.0       & 77.2          \\ \hline

\end{tabular}
\end{table}

\textbf{Single-scale vs. multi-scale attention: }
In Table~\ref{tab:tab7}, we compare the single-scale attention with two reduction ratios and multi-scale attention. For the single-scale, the smaller reduction ratio keeps more detailed information, thus contributing to better accuracy, especially for dense prediction tasks (+0.6 points for segmentation). 
By contrast, the multi-scale attention outperforms the single-scale attention by at least 0.5 points on ImageNet and at least 0.8 points on segmentation, though it suffers from a slight increase of parameters (+0.9M) and FLOPs (+0.1G). 

\begin{table}[t!]
\small
\caption{Comparison of single-scale and multi-scale attention on the ImageNet-1K and ADE-20K datasets.}
\label{tab:tab7}
\centering
\begin{tabular}{c|cccc|c|c|c|c}
\hline
\multirow{2}{*}{} & \multicolumn{4}{c|}{Reduction ratios}                                                                     & \multirow{2}{*}{Params. (M)} & \multirow{2}{*}{FLOPs (G)} & \multirow{2}{*}{Top1}  & \multirow{2}{*}{mIOU} \\ \cline{2-5}
                        & \multicolumn{1}{c|}{Stage1}     & \multicolumn{1}{c|}{Stage2}    & \multicolumn{1}{c|}{Stage3}   & Stage4 &                             &                            &                            \\ \hline
\multirow{2}{*}{Single-scale} & \multicolumn{1}{c|}{64}         & \multicolumn{1}{c|}{16}        & \multicolumn{1}{c|}{4}        & 1      &       10.8                      & 2.0                           &  77.2             & 41.2                  \\ \cline{2-9} 
                        & \multicolumn{1}{c|}{16}         & \multicolumn{1}{c|}{4}         & \multicolumn{1}{c|}{1}        & 1      &   10.8                         &  2.1                          &  77.4           & 41.8                    \\ \hline
Multi-scale                   & \multicolumn{1}{c|}{\{64, 16\}} & \multicolumn{1}{c|}{\{16, 4\}} & \multicolumn{1}{c|}{\{4, 1\}} & 1      &   11.7                          &  2.2                          & 77.9       &  42.6                  \\ \hline
\end{tabular}
\end{table}

\section{Conclusion}
The dense self-attention in Transformers suffers from the high computation complexity when processing vision tasks, especially on dense prediction scenarios or high-resolution images. 
In this work, we propose the content-based sparse attention that clusters vision tokens and aggregates them in the same cluster. The clustering-guided self-attention not only reduces the computation complexity but also invests the explicit and intensive semantics to each aggregated token, thus contributing to better performance. 
Moreover, we extend it from single-scale to multi-scale self-attention, benefiting the dense prediction tasks. 
Based on the proposed self-attention method, we build a versatile Transformer model, called ClusTR. 
We conduct extensive experiments to demonstrate the effectiveness of ClusTR, and achieve state-of-the-art performance on various vision tasks, including image recognition, semantic segmentation, object detection, and pose estimation. 


\bibliography{iclr2023_conference}

\begin{thebibliography}{56}
\providecommand{\natexlab}[1]{#1}
\providecommand{\url}[1]{\texttt{#1}}
\expandafter\ifx\csname urlstyle\endcsname\relax
  \providecommand{\doi}[1]{doi: #1}\else
  \providecommand{\doi}{doi: \begingroup \urlstyle{rm}\Url}\fi

\bibitem[Beltagy et~al.(2020)Beltagy, Peters, and Cohan]{Beltagy2020Longformer}
Iz~Beltagy, Matthew~E. Peters, and Arman Cohan.
\newblock Longformer: The long-document transformer.
\newblock \emph{arXiv:2004.05150}, 2020.

\bibitem[Brown et~al.(2020)Brown, Mann, Ryder, Subbiah, Kaplan, Dhariwal,
  Neelakantan, Shyam, Sastry, Askell, et~al.]{GPT3}
Tom Brown, Benjamin Mann, Nick Ryder, Melanie Subbiah, Jared~D Kaplan, Prafulla
  Dhariwal, Arvind Neelakantan, Pranav Shyam, Girish Sastry, Amanda Askell,
  et~al.
\newblock Language models are few-shot learners.
\newblock \emph{Advances in neural information processing systems},
  33:\penalty0 1877--1901, 2020.

\bibitem[Carion et~al.(2020)Carion, Massa, Synnaeve, Usunier, Kirillov, and
  Zagoruyko]{DETR}
Nicolas Carion, Francisco Massa, Gabriel Synnaeve, Nicolas Usunier, Alexander
  Kirillov, and Sergey Zagoruyko.
\newblock End-to-end object detection with transformers.
\newblock In \emph{European conference on computer vision}, pp.\  213--229.
  Springer, 2020.

\bibitem[Chen et~al.(2022{\natexlab{a}})Chen, Panda, and
  Fan]{chen2022regionvit}
Chun-Fu Chen, Rameswar Panda, and Quanfu Fan.
\newblock Regionvit: Regional-to-local attention for vision transformers.
\newblock In \emph{International Conference on Learning Representations},
  2022{\natexlab{a}}.

\bibitem[Chen et~al.(2021{\natexlab{a}})Chen, Fan, and Panda]{chen2021crossvit}
Chun-Fu~Richard Chen, Quanfu Fan, and Rameswar Panda.
\newblock Crossvit: Cross-attention multi-scale vision transformer for image
  classification.
\newblock In \emph{Proceedings of the IEEE/CVF international conference on
  computer vision}, pp.\  357--366, 2021{\natexlab{a}}.

\bibitem[Chen et~al.(2021{\natexlab{b}})Chen, Wang, Guo, Xu, Deng, Liu, Ma, Xu,
  Xu, and Gao]{chen2021pre}
Hanting Chen, Yunhe Wang, Tianyu Guo, Chang Xu, Yiping Deng, Zhenhua Liu, Siwei
  Ma, Chunjing Xu, Chao Xu, and Wen Gao.
\newblock Pre-trained image processing transformer.
\newblock In \emph{Proceedings of the IEEE/CVF Conference on Computer Vision
  and Pattern Recognition}, pp.\  12299--12310, 2021{\natexlab{b}}.

\bibitem[Chen et~al.(2019)Chen, Wang, Pang, et~al.]{chen2019mmdetection}
Kai Chen, Jiaqi Wang, Jiangmiao Pang, et~al.
\newblock Mmdetection: Open mmlab detection toolbox and benchmark.
\newblock \emph{arXiv preprint arXiv:1906.07155}, 2019.

\bibitem[Chen et~al.(2022{\natexlab{b}})Chen, Xie, Chongjian, Chen, Liang, and
  Luo]{chen2022cyclemlp}
Shoufa Chen, Enze Xie, GE~Chongjian, Runjian Chen, Ding Liang, and Ping Luo.
\newblock Cyclemlp: A mlp-like architecture for dense prediction.
\newblock In \emph{International Conference on Learning Representations},
  2022{\natexlab{b}}.

\bibitem[Chu et~al.(2021)Chu, Tian, Wang, Zhang, Ren, Wei, Xia, and
  Shen]{chu2021twins}
Xiangxiang Chu, Zhi Tian, Yuqing Wang, Bo~Zhang, Haibing Ren, Xiaolin Wei,
  Huaxia Xia, and Chunhua Shen.
\newblock Twins: Revisiting the design of spatial attention in vision
  transformers.
\newblock \emph{Advances in Neural Information Processing Systems},
  34:\penalty0 9355--9366, 2021.

\bibitem[Deng et~al.(2009)Deng, Dong, Socher, Li, Li, and
  Fei-Fei]{deng2009imagenet}
Jia Deng, Wei Dong, Richard Socher, Li-Jia Li, Kai Li, and Li~Fei-Fei.
\newblock Imagenet: A large-scale hierarchical image database.
\newblock In \emph{2009 IEEE conference on computer vision and pattern
  recognition}, pp.\  248--255. Ieee, 2009.

\bibitem[Dong et~al.(2022)Dong, Bao, Chen, Zhang, Yu, Yuan, Chen, and
  Guo]{dong2022cswin}
Xiaoyi Dong, Jianmin Bao, Dongdong Chen, Weiming Zhang, Nenghai Yu, Lu~Yuan,
  Dong Chen, and Baining Guo.
\newblock Cswin transformer: A general vision transformer backbone with
  cross-shaped windows.
\newblock In \emph{Proceedings of the IEEE/CVF Conference on Computer Vision
  and Pattern Recognition}, pp.\  12124--12134, 2022.

\bibitem[Dosovitskiy et~al.(2021)Dosovitskiy, Beyer, Kolesnikov, Weissenborn,
  Zhai, Unterthiner, Dehghani, Minderer, Heigold, Gelly, et~al.]{ViT}
Alexey Dosovitskiy, Lucas Beyer, Alexander Kolesnikov, Dirk Weissenborn,
  Xiaohua Zhai, Thomas Unterthiner, Mostafa Dehghani, Matthias Minderer, Georg
  Heigold, Sylvain Gelly, et~al.
\newblock An image is worth 16x16 words: Transformers for image recognition at
  scale.
\newblock In \emph{International Conference on Learning Representations}, 2021.

\bibitem[d’Ascoli et~al.(2021)d’Ascoli, Touvron, Leavitt, Morcos, Biroli,
  and Sagun]{d2021convit}
St{\'e}phane d’Ascoli, Hugo Touvron, Matthew~L Leavitt, Ari~S Morcos, Giulio
  Biroli, and Levent Sagun.
\newblock Convit: Improving vision transformers with soft convolutional
  inductive biases.
\newblock In \emph{International Conference on Machine Learning}, pp.\
  2286--2296. PMLR, 2021.

\bibitem[Gu et~al.(2022)Gu, Kwon, Wang, Ye, Li, Chen, Lai, Chandra, and
  Pan]{gu2022multi}
Jiaqi Gu, Hyoukjun Kwon, Dilin Wang, Wei Ye, Meng Li, Yu-Hsin Chen, Liangzhen
  Lai, Vikas Chandra, and David~Z Pan.
\newblock Multi-scale high-resolution vision transformer for semantic
  segmentation.
\newblock In \emph{Proceedings of the IEEE/CVF Conference on Computer Vision
  and Pattern Recognition}, pp.\  12094--12103, 2022.

\bibitem[He et~al.(2016)He, Zhang, Ren, and Sun]{he2016deep}
Kaiming He, Xiangyu Zhang, Shaoqing Ren, and Jian Sun.
\newblock Deep residual learning for image recognition.
\newblock In \emph{Proceedings of the IEEE conference on computer vision and
  pattern recognition}, pp.\  770--778, 2016.

\bibitem[He et~al.(2017)He, Gkioxari, Doll{\'a}r, and Girshick]{he2017mask}
Kaiming He, Georgia Gkioxari, Piotr Doll{\'a}r, and Ross Girshick.
\newblock Mask r-cnn.
\newblock In \emph{Proceedings of the IEEE international conference on computer
  vision}, pp.\  2961--2969, 2017.

\bibitem[Heo et~al.(2021)Heo, Yun, Han, Chun, Choe, and Oh]{PiT}
Byeongho Heo, Sangdoo Yun, Dongyoon Han, Sanghyuk Chun, Junsuk Choe, and
  Seong~Joon Oh.
\newblock Rethinking spatial dimensions of vision transformers.
\newblock In \emph{Proceedings of the IEEE/CVF International Conference on
  Computer Vision}, pp.\  11936--11945, 2021.

\bibitem[Jiang et~al.(2021)Jiang, Chang, and Wang]{jiang2021transgan}
Yifan Jiang, Shiyu Chang, and Zhangyang Wang.
\newblock Transgan: Two pure transformers can make one strong gan, and that can
  scale up.
\newblock \emph{Advances in Neural Information Processing Systems}, 34, 2021.

\bibitem[Jin et~al.(2020)Jin, Xu, Xu, Wang, Liu, Qian, Ouyang, and
  Luo]{jin2020whole}
Sheng Jin, Lumin Xu, Jin Xu, Can Wang, Wentao Liu, Chen Qian, Wanli Ouyang, and
  Ping Luo.
\newblock Whole-body human pose estimation in the wild.
\newblock In \emph{European Conference on Computer Vision}, pp.\  196--214.
  Springer, 2020.

\bibitem[Kirillov et~al.(2019)Kirillov, Girshick, He, and Dollar]{2019Panoptic}
A.~Kirillov, R.~Girshick, K.~He, and P.~Dollar.
\newblock Panoptic feature pyramid networks.
\newblock In \emph{2019 IEEE/CVF Conference on Computer Vision and Pattern
  Recognition (CVPR)}, 2019.

\bibitem[Kitaev et~al.(2019)Kitaev, Kaiser, and Levskaya]{Reformer}
Nikita Kitaev, Lukasz Kaiser, and Anselm Levskaya.
\newblock Reformer: The efficient transformer.
\newblock In \emph{International Conference on Learning Representations}, 2019.

\bibitem[Krizhevsky et~al.(2012)Krizhevsky, Sutskever, and Hinton]{AlexNet}
Alex Krizhevsky, Ilya Sutskever, and Geoffrey~E Hinton.
\newblock Imagenet classification with deep convolutional neural networks.
\newblock \emph{Advances in neural information processing systems}, 25, 2012.

\bibitem[Lee et~al.(2022)Lee, Kim, Willette, and Hwang]{Lee_2022_CVPR}
Youngwan Lee, Jonghee Kim, Jeffrey Willette, and Sung~Ju Hwang.
\newblock Mpvit: Multi-path vision transformer for dense prediction.
\newblock In \emph{Proceedings of the IEEE/CVF Conference on Computer Vision
  and Pattern Recognition (CVPR)}, pp.\  7287--7296, June 2022.

\bibitem[Li et~al.(2021)Li, Zhang, Cao, Timofte, and Van~Gool]{LocalViT}
Yawei Li, Kai Zhang, Jiezhang Cao, Radu Timofte, and Luc Van~Gool.
\newblock Localvit: Bringing locality to vision transformers.
\newblock \emph{arXiv preprint arXiv:2104.05707}, 2021.

\bibitem[Lian et~al.(2022)Lian, Yu, Sun, and Gao]{lian2022mlp}
Dongze Lian, Zehao Yu, Xing Sun, and Shenghua Gao.
\newblock As-mlp: An axial shifted mlp architecture for vision.
\newblock In \emph{International Conference on Learning Representations}, 2022.

\bibitem[Lin et~al.(2014)Lin, Maire, Belongie, Hays, Perona, Ramanan,
  Doll{\'a}r, and Zitnick]{lin2014microsoft}
Tsung-Yi Lin, Michael Maire, Serge Belongie, James Hays, Pietro Perona, Deva
  Ramanan, Piotr Doll{\'a}r, and C~Lawrence Zitnick.
\newblock Microsoft coco: Common objects in context.
\newblock In \emph{European conference on computer vision}, pp.\  740--755.
  Springer, 2014.

\bibitem[Lin et~al.(2017)Lin, Goyal, Girshick, He, and
  Doll{\'a}r]{lin2017focal}
Tsung-Yi Lin, Priya Goyal, Ross Girshick, Kaiming He, and Piotr Doll{\'a}r.
\newblock Focal loss for dense object detection.
\newblock In \emph{Proceedings of the IEEE international conference on computer
  vision}, pp.\  2980--2988, 2017.

\bibitem[Liu et~al.(2021)Liu, Lin, Cao, Hu, Wei, Zhang, Lin, and Guo]{Swin}
Ze~Liu, Yutong Lin, Yue Cao, Han Hu, Yixuan Wei, Zheng Zhang, Stephen Lin, and
  Baining Guo.
\newblock Swin transformer: Hierarchical vision transformer using shifted
  windows.
\newblock In \emph{Proceedings of the IEEE/CVF International Conference on
  Computer Vision}, pp.\  10012--10022, 2021.

\bibitem[Liu et~al.(2022)Liu, Mao, Wu, Feichtenhofer, Darrell, and
  Xie]{liu2022convnet}
Zhuang Liu, Hanzi Mao, Chao-Yuan Wu, Christoph Feichtenhofer, Trevor Darrell,
  and Saining Xie.
\newblock A convnet for the 2020s.
\newblock In \emph{Proceedings of the IEEE/CVF Conference on Computer Vision
  and Pattern Recognition}, pp.\  11976--11986, 2022.

\bibitem[Loshchilov \& Hutter(2017)Loshchilov and Hutter]{cosine_LR}
Ilya Loshchilov and Frank Hutter.
\newblock Sgdr: Stochastic gradient descent with warm restarts.
\newblock In \emph{ICLR}, 2017.

\bibitem[Loshchilov \& Hutter(2018)Loshchilov and Hutter]{adamw}
Ilya Loshchilov and Frank Hutter.
\newblock Fixing weight decay regularization in adam.
\newblock 2018.

\bibitem[Pan et~al.(2022)Pan, Ge, Lu, Song, Chen, Huang, and
  Huang]{pan2022integration}
Xuran Pan, Chunjiang Ge, Rui Lu, Shiji Song, Guanfu Chen, Zeyi Huang, and Gao
  Huang.
\newblock On the integration of self-attention and convolution.
\newblock In \emph{Proceedings of the IEEE/CVF Conference on Computer Vision
  and Pattern Recognition}, pp.\  815--825, 2022.

\bibitem[Radosavovic et~al.(2020)Radosavovic, Kosaraju, Girshick, He, and
  Doll{\'a}r]{radosavovic2020designing}
Ilija Radosavovic, Raj~Prateek Kosaraju, Ross Girshick, Kaiming He, and Piotr
  Doll{\'a}r.
\newblock Designing network design spaces.
\newblock In \emph{Proceedings of the IEEE/CVF conference on computer vision
  and pattern recognition}, pp.\  10428--10436, 2020.

\bibitem[Ren et~al.(2022)Ren, Zhou, He, Feng, and Wang]{ren2022shunted}
Sucheng Ren, Daquan Zhou, Shengfeng He, Jiashi Feng, and Xinchao Wang.
\newblock Shunted self-attention via multi-scale token aggregation.
\newblock In \emph{Proceedings of the IEEE/CVF Conference on Computer Vision
  and Pattern Recognition}, pp.\  10853--10862, 2022.

\bibitem[Rodriguez \& Laio(2014)Rodriguez and Laio]{rodriguez2014clustering}
Alex Rodriguez and Alessandro Laio.
\newblock Clustering by fast search and find of density peaks.
\newblock \emph{science}, 344\penalty0 (6191):\penalty0 1492--1496, 2014.

\bibitem[Roy et~al.(2021)Roy, Saffar, Vaswani, and
  Grangier]{routingTransformer}
Aurko Roy, Mohammad Saffar, Ashish Vaswani, and David Grangier.
\newblock Efficient content-based sparse attention with routing transformers.
\newblock \emph{Transactions of the Association for Computational Linguistics},
  9:\penalty0 53--68, 2021.

\bibitem[Sun et~al.(2019)Sun, Xiao, Liu, and Wang]{sun2019deep}
Ke~Sun, Bin Xiao, Dong Liu, and Jingdong Wang.
\newblock Deep high-resolution representation learning for human pose
  estimation.
\newblock In \emph{Proceedings of the IEEE/CVF conference on computer vision
  and pattern recognition}, pp.\  5693--5703, 2019.

\bibitem[Szegedy et~al.(2015)Szegedy, Liu, Jia, Sermanet, Reed, Anguelov,
  Erhan, Vanhoucke, and Rabinovich]{GoogleNet}
Christian Szegedy, Wei Liu, Yangqing Jia, Pierre Sermanet, Scott Reed, Dragomir
  Anguelov, Dumitru Erhan, Vincent Vanhoucke, and Andrew Rabinovich.
\newblock Going deeper with convolutions.
\newblock In \emph{Proceedings of the IEEE conference on computer vision and
  pattern recognition}, pp.\  1--9, 2015.

\bibitem[Tan \& Le(2019)Tan and Le]{tan2019efficientnet}
Mingxing Tan and Quoc Le.
\newblock Efficientnet: Rethinking model scaling for convolutional neural
  networks.
\newblock In \emph{International conference on machine learning}, pp.\
  6105--6114. PMLR, 2019.

\bibitem[Tang et~al.(2022)Tang, Zhang, Zhu, and Tan]{tang2022quadtree}
Shitao Tang, Jiahui Zhang, Siyu Zhu, and Ping Tan.
\newblock Quadtree attention for vision transformers.
\newblock In \emph{International Conference on Learning Representations}, 2022.

\bibitem[Touvron et~al.(2021{\natexlab{a}})Touvron, Cord, Douze, Massa,
  Sablayrolles, and J{\'e}gou]{DeiT}
Hugo Touvron, Matthieu Cord, Matthijs Douze, Francisco Massa, Alexandre
  Sablayrolles, and Herv{\'e} J{\'e}gou.
\newblock Training data-efficient image transformers \& distillation through
  attention.
\newblock In \emph{International Conference on Machine Learning}, pp.\
  10347--10357. PMLR, 2021{\natexlab{a}}.

\bibitem[Touvron et~al.(2021{\natexlab{b}})Touvron, Cord, Sablayrolles,
  Synnaeve, and J{\'e}gou]{CaiT}
Hugo Touvron, Matthieu Cord, Alexandre Sablayrolles, Gabriel Synnaeve, and
  Herv{\'e} J{\'e}gou.
\newblock Going deeper with image transformers.
\newblock In \emph{Proceedings of the IEEE/CVF International Conference on
  Computer Vision}, pp.\  32--42, 2021{\natexlab{b}}.

\bibitem[Vaswani et~al.(2017)Vaswani, Shazeer, Parmar, Uszkoreit, Jones, Gomez,
  Kaiser, and Polosukhin]{vaswani2017attention}
Ashish Vaswani, Noam Shazeer, Niki Parmar, Jakob Uszkoreit, Llion Jones,
  Aidan~N Gomez, {\L}ukasz Kaiser, and Illia Polosukhin.
\newblock Attention is all you need.
\newblock \emph{Advances in neural information processing systems}, 30, 2017.

\bibitem[Wang et~al.(2022{\natexlab{a}})Wang, Wang, Wang, Lin, Chang, Xie, Li,
  and Jin]{wang2022kvt}
Pichao Wang, Xue Wang, Fan Wang, Ming Lin, Shuning Chang, Wen Xie, Hao Li, and
  Rong Jin.
\newblock Kvt: k-nn attention for boosting vision transformers.
\newblock In \emph{European conference on computer vision}, 2022{\natexlab{a}}.

\bibitem[Wang et~al.(2021)Wang, Xie, Li, Fan, Song, Liang, Lu, Luo, and
  Shao]{PVT}
Wenhai Wang, Enze Xie, Xiang Li, Deng-Ping Fan, Kaitao Song, Ding Liang, Tong
  Lu, Ping Luo, and Ling Shao.
\newblock Pyramid vision transformer: A versatile backbone for dense prediction
  without convolutions.
\newblock In \emph{Proceedings of the IEEE/CVF International Conference on
  Computer Vision (ICCV)}, pp.\  568--578, October 2021.

\bibitem[Wang et~al.(2022{\natexlab{b}})Wang, Xie, Li, Fan, Song, Liang, Lu,
  Luo, and Shao]{wang2021pvtv2}
Wenhai Wang, Enze Xie, Xiang Li, Deng-Ping Fan, Kaitao Song, Ding Liang, Tong
  Lu, Ping Luo, and Ling Shao.
\newblock Pvtv2: Improved baselines with pyramid vision transformer.
\newblock \emph{Computational Visual Media}, 8\penalty0 (3):\penalty0 1--10,
  2022{\natexlab{b}}.

\bibitem[Wang et~al.(2022{\natexlab{c}})Wang, Yao, Chen, Lin, Cai, He, and
  Liu]{wang2022crossformer}
Wenxiao Wang, Lu~Yao, Long Chen, Binbin Lin, Deng Cai, Xiaofei He, and Wei Liu.
\newblock Crossformer: A versatile vision transformer hinging on cross-scale
  attention.
\newblock In \emph{International Conference on Learning Representations},
  2022{\natexlab{c}}.

\bibitem[Xiao et~al.(2018{\natexlab{a}})Xiao, Wu, and Wei]{xiao2018simple}
Bin Xiao, Haiping Wu, and Yichen Wei.
\newblock Simple baselines for human pose estimation and tracking.
\newblock In \emph{Proceedings of the European conference on computer vision
  (ECCV)}, pp.\  466--481, 2018{\natexlab{a}}.

\bibitem[Xiao et~al.(2018{\natexlab{b}})Xiao, Liu, Zhou, Jiang, and
  Sun]{xiao2018unified}
Tete Xiao, Yingcheng Liu, Bolei Zhou, Yuning Jiang, and Jian Sun.
\newblock Unified perceptual parsing for scene understanding.
\newblock In \emph{European Conference on Computer Vision}. Springer,
  2018{\natexlab{b}}.

\bibitem[Yuan et~al.(2021)Yuan, Chen, Wang, Yu, Shi, Jiang, Tay, Feng, and
  Yan]{T2T-ViT}
Li~Yuan, Yunpeng Chen, Tao Wang, Weihao Yu, Yujun Shi, Zi-Hang Jiang,
  Francis~EH Tay, Jiashi Feng, and Shuicheng Yan.
\newblock Tokens-to-token vit: Training vision transformers from scratch on
  imagenet.
\newblock In \emph{Proceedings of the IEEE/CVF International Conference on
  Computer Vision}, pp.\  558--567, 2021.

\bibitem[Zaheer et~al.(2020)Zaheer, Guruganesh, Dubey, Ainslie, Alberti,
  Ontanon, Pham, Ravula, Wang, Yang, et~al.]{zaheer2020big}
Manzil Zaheer, Guru Guruganesh, Kumar~Avinava Dubey, Joshua Ainslie, Chris
  Alberti, Santiago Ontanon, Philip Pham, Anirudh Ravula, Qifan Wang, Li~Yang,
  et~al.
\newblock Big bird: Transformers for longer sequences.
\newblock \emph{Advances in Neural Information Processing Systems},
  33:\penalty0 17283--17297, 2020.

\bibitem[Zeng et~al.(2022{\natexlab{a}})Zeng, Jin, Liu, Qian, Luo, Ouyang, and
  Wang]{TCFormer}
Wang Zeng, Sheng Jin, Wentao Liu, Chen Qian, Ping Luo, Wanli Ouyang, and
  Xiaogang Wang.
\newblock Not all tokens are equal: Human-centric visual analysis via token
  clustering transformer.
\newblock In \emph{Proceedings of the IEEE/CVF Conference on Computer Vision
  and Pattern Recognition}, pp.\  11101--11111, 2022{\natexlab{a}}.

\bibitem[Zeng et~al.(2022{\natexlab{b}})Zeng, Jin, Liu, Qian, Luo, Ouyang, and
  Wang]{zeng2022not}
Wang Zeng, Sheng Jin, Wentao Liu, Chen Qian, Ping Luo, Wanli Ouyang, and
  Xiaogang Wang.
\newblock Not all tokens are equal: Human-centric visual analysis via token
  clustering transformer.
\newblock In \emph{Proceedings of the IEEE/CVF Conference on Computer Vision
  and Pattern Recognition}, pp.\  11101--11111, 2022{\natexlab{b}}.

\bibitem[Zhang et~al.(2021)Zhang, Dai, Yang, Xiao, Yuan, Zhang, and
  Gao]{zhang2021multi}
Pengchuan Zhang, Xiyang Dai, Jianwei Yang, Bin Xiao, Lu~Yuan, Lei Zhang, and
  Jianfeng Gao.
\newblock Multi-scale vision longformer: A new vision transformer for
  high-resolution image encoding.
\newblock In \emph{Proceedings of the IEEE/CVF International Conference on
  Computer Vision}, pp.\  2998--3008, 2021.

\bibitem[Zheng et~al.(2021)Zheng, Lu, Zhao, Zhu, Luo, Wang, Fu, Feng, Xiang,
  Torr, et~al.]{SETR}
Sixiao Zheng, Jiachen Lu, Hengshuang Zhao, Xiatian Zhu, Zekun Luo, Yabiao Wang,
  Yanwei Fu, Jianfeng Feng, Tao Xiang, Philip~HS Torr, et~al.
\newblock Rethinking semantic segmentation from a sequence-to-sequence
  perspective with transformers.
\newblock In \emph{Proceedings of the IEEE/CVF conference on computer vision
  and pattern recognition}, pp.\  6881--6890, 2021.

\bibitem[Zhou et~al.(2017)Zhou, Zhao, Puig, Fidler, Barriuso, and
  Torralba]{zhou2017scene}
Bolei Zhou, Hang Zhao, Xavier Puig, Sanja Fidler, Adela Barriuso, and Antonio
  Torralba.
\newblock Scene parsing through ade20k dataset.
\newblock In \emph{Proceedings of the IEEE conference on computer vision and
  pattern recognition}, pp.\  633--641, 2017.

\end{thebibliography}
\bibliographystyle{iclr2023_conference}

\end{document}